\documentclass{llncs}
\usepackage{llncsdoc}

\usepackage{amsmath}
\usepackage{amsfonts}
\usepackage[dvipsnames]{xcolor}
\usepackage{graphicx}
\usepackage{algorithm}
\usepackage{algpseudocode}
\usepackage{multirow}
\usepackage{hyperref}
\usepackage{listings}

\lstdefinestyle{custompython}{
  belowcaptionskip=1\baselineskip,
  breaklines=true,
  frame=L,
  language=Python,
  showstringspaces=false,
  basicstyle=\fontsize{8}{10}\selectfont\ttfamily,
  keywordstyle=\bfseries\color{blue!60!black},
  identifierstyle=\color{black},
  stringstyle=\color{red!50!black},
}
\DeclareMathOperator*{\argmax}{arg\,max}

\title{A Comprehensive Implementation of Conceptual Spaces\thanks{The final paper is published at \url{http://ceur-ws.org/Vol-2090/}}}
\author{Lucas Bechberger\thanks{Corresponding author, ORCID: 0000-0002-1962-1777} \and Kai-Uwe K\"uhnberger}
\institute{Institute of Cognitive Science, Osnabr\"uck University, Osnabr\"uck, Germany \email{lucas.bechberger@uni-osnabrueck.de}, \email{kai-uwe.kuehnberger@uni-osnabrueck.de}}

\begin{document}
\maketitle

\begin{abstract}
The highly influential framework of conceptual spaces provides a geometric way of representing knowledge. Instances are represented by points and concepts are represented by regions in a (potentially) high-dimensional space. Based on our recent formalization, we present a comprehensive implementation of the conceptual spaces framework that is not only capable of representing concepts with inter-domain correlations, but that also offers a variety of operations on these concepts.
\end{abstract}
\section{Introduction}
\label{Intro}

One common criticism of symbolic AI approaches is that the symbols they operate on do not contain any meaning: For the system, they are just arbitrary tokens that can be manipulated in some way. This lack of inherent meaning in abstract symbols is called the “symbol grounding problem” \cite{Harnad1990}. One approach towards solving this problem is to devise a grounding mechanism that connects  abstract symbols to the real world, i.e., to perception and action.

The cognitive framework of conceptual spaces \cite{Gardenfors2000,Gardenfors2014} attempts to bridge this gap between symbolic and subsymbolic AI by proposing an intermediate conceptual layer based on geometric representations.
A conceptual space is spanned by a number of quality dimensions that are based on perception and/or subsymbolic processing. Regions in this space correspond to concepts and can be referred to as abstract symbols.

The framework of conceptual spaces has been highly influential in the last 15 years within cognitive science and cognitive linguistics \cite{Douven2011,Fiorini2013,Warglien2012}. It has also sparked considerable research in various subfields of artificial intelligence, ranging from robotics and computer vision \cite{Chella2001,Chella2003} over the semantic web \cite{Adams2009a} to plausible reasoning \cite{Derrac2015,Schockaert2011}.\\

Most practical implementations of the conceptual spaces framework are rather ad-hoc and tailored towards a specific application. They tend to ignore important aspects of the framework and should thus be regarded as only partial implementations of the framework. Furthermore, these implementations are usually not publicly accessible, which greatly reduces their value for the research community.

In this paper, we present a thorough and comprehensive implementation of the conceptual spaces framework which is based on our formalization reported in \cite{Bechberger2017KI} and \cite{Bechberger2017SGAI}. Its source code is publicly available on GitHub to researchers anywhere in the world. Instead of investing a considerable amount of time into developing their own implementation, researchers can use our implementation off-the-shelf and focus on their specific application scenario.

The remainder of this paper is structured as follows:
Section \ref{CS} introduces the framework of conceptual spaces and our formalization. In Section \ref{Implementation}, we give an overview of our implementation and in Section \ref{Examples} we illustrate its usage. Section \ref{RelatedWork} summarizes related work and Section \ref{Conclusion} concludes the paper.

\section{Conceptual Spaces}
\label{CS}

This section presents the cognitive framework of conceptual spaces as described in \cite{Gardenfors2000} and summarizes our formalization as reported in \cite{Bechberger2017KI} and \cite{Bechberger2017SGAI}.

\subsection{Dimensions, Domains, and Distance}
\label{CS:DimensionsDomainsDistance}

A conceptual space is spanned by a set $D$ of so-called ``quality dimensions''. Each of these dimensions $d \in D$ represents a way in which two stimuli can be judged to be similar or different. Examples for quality dimensions include temperature, weight, time, pitch, and hue. The distance between two points $x$ and $y$ with respect to a dimension $d$ is denoted as $|x_d - y_d|$.

A domain $\delta \subseteq D$ is a set of dimensions that inherently belong together. Different perceptual modalities (like color, shape, or taste) are represented by different domains. The color domain for instance consists of the three dimensions hue, saturation, and brightness. Distance within a domain $\delta$ is measured by the weighted Euclidean metric $d_E$.  

The overall conceptual space $CS$ is defined as the product space of all dimensions. Distance within the overall conceptual space is measured by the weighted Manhattan metric $d_M$ of the intra-domain distances. This is supported by both psychological evidence \cite{Attneave1950,Shepard1964} and mathematical considerations \cite{Aggarwal2001}. Let $\Delta$ be the set of all domains in $CS$. The combined distance $d_C^\Delta$ within $CS$ is defined as follows:
$$
d_C^{\Delta}(x,y,W) = \sum_{\delta \in \Delta}w_{\delta} \cdot \sqrt{\sum_{d \in \delta} w_{d} \cdot |x_{d} - y_{d}|^2}
$$
The parameter $W = \langle W_{\Delta},\{W_{\delta}\}_{\delta \in \Delta}\rangle$ contains two parts: $W_{\Delta}$ is the set of positive domain weights $w_{\delta}$ with $\textstyle\sum_{\delta \in \Delta} w_{\delta} = |\Delta|$. Moreover, $W$ contains for each domain $\delta \in \Delta$ a set $W_{\delta}$ of positive dimension weights $w_{d}$ with $\textstyle\sum_{d \in \delta} w_{d} = 1$.\\

The similarity of two points in a conceptual space is inversely related to their distance. This can be written as follows :
$$Sim(x,y) = e^{-c \cdot d(x,y)}\quad \text{with a constant}\; c >0 \; \text{and a given metric}\; d$$

Betweenness is a logical predicate $B(x,y,z)$ that is true if and only if $y$ is considered to be between $x$ and $z$. It can be defined based on a given metric $d$: 
$$B_d(x,y,z) :\iff d(x,y) + d(y,z) = d(x,z)$$

The betweenness relation based on $d_E$ results in the line segment connecting the points $x$ and $z$, whereas the betweenness relation based on $d_M$ results in an axis-parallel cuboid between the points $x$ and $z$.
One can define convexity and star-shapedness based on the notion of betweenness:

\begin{definition}
\label{def:Convexity}
(Convexity)\\
A set $C \subseteq CS$ is \emph{convex} under a metric $d \;:\iff$

\hspace{1cm}$\forall {x \in C, z \in C, y \in CS}: \left(B_d(x,y,z) \rightarrow y \in C\right)$
\end{definition}

\begin{definition}
\label{def:StarShapedSet}
(Star-shapedness)\\
A set $S \subseteq CS$ is \emph{star-shaped} under a metric $d$ with respect to a set $P \subseteq S \;:\iff$ 

\hspace{1cm}$\forall {p \in P, z \in S, y \in CS}: \left(B_d(p,y,z) \rightarrow y \in S\right)$
\end{definition}

\subsection{Properties and Concepts}
\label{CS:PropertiesConcepts}

G\"{a}rdenfors \cite{Gardenfors2000} distinguishes properties like ``red'', ``round'', and ``sweet'' from full-fleshed concepts like ``apple'' or ``dog'' by observing that properties can be defined on individual domains (e.g., color, shape, taste), whereas full-fleshed concepts involve multiple domains.
Each domain involved in representing a concept has a certain importance, which is reflected by so-called ``salience weights''. Another important aspect of concepts are the correlations between the different domains, which are important for both learning \cite{Billman1996} and reasoning \cite[Ch 8]{Murphy2002}.

\begin{figure}[tp]
\centering
\includegraphics[width=0.7\columnwidth]{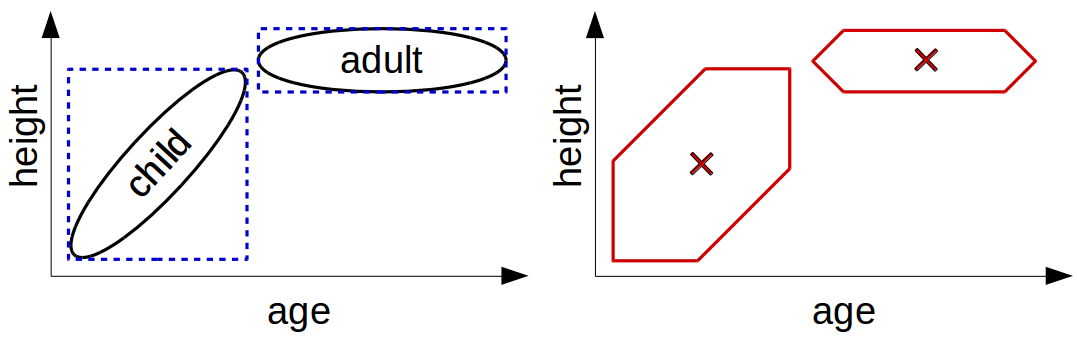}
\caption{Left: Intuitive way to define regions for the concepts of ``adult'' and ``child'' (solid) as well as representation by using convex sets (dashed). Right: Representation by using star-shaped sets with central points marked by crosses.}
\label{fig:ConvexityProblem}
\end{figure}

Based on the principle of cognitive economy, G\"{a}rdenfors argues that both properties and concepts should be represented as convex sets. However, this convexity assumption has recently been criticized \cite{Hernandez-Conde2016} and we demonstrated in \cite{Bechberger2017KI} that one cannot geometrically encode correlations between domains when using convex sets:
The left part of Figure \ref{fig:ConvexityProblem} shows two domains, age and height, and the concepts of child and adult. The solid ellipses illustrate the intuitive way of defining these concepts. As domains are combined with the Manhattan metric, a convex set corresponds in this case to an axis-parallel cuboid. One can easily see that this convex representation (dashed rectangles) is not satisfactory, because the correlation of the two domains is not encoded. We therefore proposed in \cite{Bechberger2017KI} to relax the convexity criterion and to use star-shaped sets, which is illustrated in the right part of Figure \ref{fig:ConvexityProblem}. This enables a geometric representation of correlations while still being only a minimal departure from the original framework.\\

We have based our formalization on axis-parallel cuboids that can be described by a triple $\langle \Delta_C, p^-, p^+ \rangle$ consisting of a set of domains $\Delta_C$ on which this cuboid $C$ is defined and two points $p^-$ and $p^+$, such that 
$$x \in C \iff \forall{\delta \in \Delta_C}:\forall{d \in \delta}: p_d^- \leq x_d \leq p_d^+$$
These cuboids are convex under $d_C^{\Delta}$. It is also easy to see that any union of convex sets that have a non-empty intersection is star-shaped \cite{Smith1968}. We define the core of a concept as follows:

\begin{definition}
\label{def:SSSS}
(Simple star-shaped set)\\
A \emph{simple star-shaped set} $S$ is described as a tuple $\langle\Delta_S,\{C_1,\dots,C_m\}\rangle$. $\Delta_S \subseteq \Delta$ is a set of domains on which the cuboids $\{C_1,\dots,C_m\}$ (and thus also $S$) are defined. Moreover, we require that the central region $P :=\textstyle\bigcap_{i = 1}^m C_i \neq \emptyset$. Then the simple star-shaped set $S$ is defined as 
$$S := \bigcup_{i=1}^m C_i$$
\end{definition}

In order to represent imprecise concept boundaries, we use fuzzy sets \cite{Bvelohlavek2011,Zadeh1965,Zadeh1982}. A fuzzy set is characterized by its membership function $\mu: CS \rightarrow [0,1]$ which assigns a degree of membership to each point in the conceptual space. The membership of a point to a fuzzy concept is based on its maximal similarity to any of the points in the concept's core:

\begin{definition}
\label{def:FSSSS}
(Fuzzy simple star-shaped set)\\
A \emph{fuzzy simple star-shaped set} $\widetilde{S}$ is described by a quadruple $\langle S,\mu_0,c,W\rangle$ where
$S = \langle\Delta_S,\{C_1,\dots,C_m\}\rangle$ is a non-empty simple star-shaped set. The parameter $\mu_0 \in (0,1]$ controls the highest possible membership to $\widetilde{S}$ and is usually set to 1. The sensitivity parameter $c > 0$ controls the rate of the exponential decay in the similarity function. Finally, $W = \langle W_{\Delta_S},\{W_{\delta}\}_{\delta \in \Delta_S}\rangle$ contains positive weights for all domains in $\Delta_S$ and all dimensions within these domains, reflecting their respective importance. We require that $\textstyle\sum_{\delta \in \Delta_S} w_{\delta} = |\Delta_S|$ and that $\forall {\delta \in \Delta_S}:\textstyle\sum_{d \in \delta} w_{d} = 1$.
The membership function of $\widetilde{S}$ is then defined as follows:
$$\mu_{\widetilde{S}}(x) = \mu_0 \cdot \max_{y \in S}(e^{-c \cdot d_C^{\Delta}(x,y,W)})$$
\end{definition}

\begin{figure}[tp]
\centering
\includegraphics[width = 0.9\columnwidth]{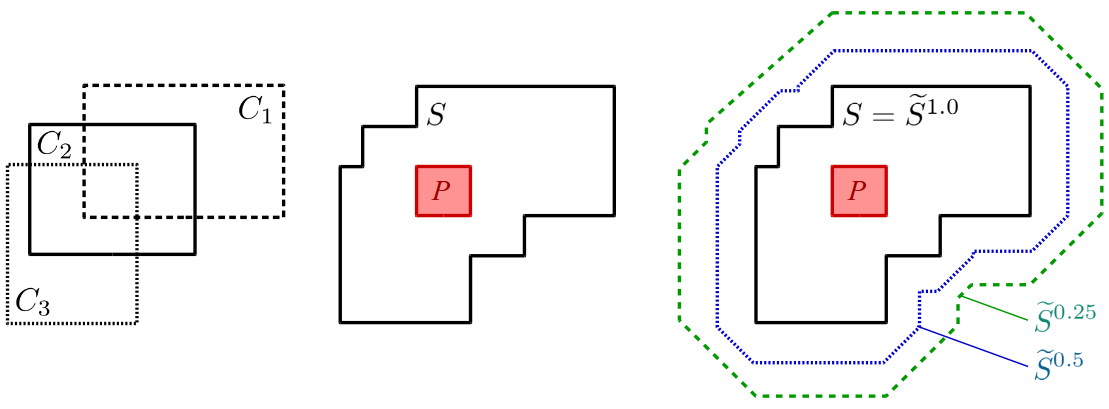} 
\caption{Left: Three cuboids $C_1, C_2, C_3$ with nonempty intersection. Middle: Resulting simple star-shaped set $S$ based on these cuboids. Right: Fuzzy simple star-shaped set $\tilde{S}$ based on $S$ with three $\alpha$-cuts for $\alpha \in \{1.0,0.5,0.25\}$.}
\label{fig:FSSSS}
\end{figure}

The sensitivity parameter $c$ controls the overall degree of fuzziness of $\widetilde{S}$ by determining how fast the membership drops to zero. The weights $W$ represent not only the relative importance of the respective domain or dimension for the represented concept, but they also influence the relative fuzziness with respect to this domain or dimension.
Note that if $|\Delta_S| = 1$, then $\widetilde{S}$ represents a property, and if $|\Delta_S| > 1$, then $\widetilde{S}$ represents a concept.
Figure \ref{fig:FSSSS} illustrates these definitions (the $x$ and $y$ axes are assumed to belong to different domains and are combined with $d_M$ using equal weights).

\subsection{Operations on Concepts}
\label{CS:Operations}

Our formalization provides a number of operations, which can be used to create new concepts from existing ones and to describe relations between concepts. We summarize them here only briefly, more details can be found in \cite{Bechberger2017KI} and \cite{Bechberger2017SGAI}.
\subsubsection{Intersection.}
\label{CS:Operations:Intersection}
The intersection of two concepts can be interpreted as the logical ``and'' -- e.g., intersecting the property ``green'' with the concept ``banana'' results in the set of all objects that are both green and bananas.
The intersection of two concepts is defined as follows:
The core of the resulting concept is the intersection of the highest intersecting $\alpha$-cuts\footnote{The $\alpha$-cut of $\widetilde{S}$ is defined as $\widetilde{S}^{\alpha} := \{ x \in CS \;|\; \mu_{\widetilde{S}}(x) \geq \alpha\}$.} of both original concepts, i.e., $S' = \widetilde{S}_1^{\alpha'} \cap \widetilde{S}_2^{\alpha'}$ with $\alpha' = \max\{\alpha \in [0,1]: {\widetilde{S}}_1^{\alpha} \cap {\widetilde{S}}_2^{\alpha} \neq \emptyset\}$. As this intersection is not guaranteed to be a valid core, we approximate it by cuboids. If these cuboids do not intersect, we compute the arithmetic mean of the cuboids' centers and extend each cuboid, such that it contains this central point. We use $\alpha'$ as the resulting concept's $\mu_0$ parameter, set $c' := \min(c^{(1)},c^{(2)})$, and derive a new set of weights by interpolating between the two original sets of weights.

\subsubsection{Union.}
\label{CS:Operations:Union}

The union of two concepts can be used to construct more abstract categories (e.g., defining ``fruit'' as the union of ``apple'', ``banana'', ``coconut'', etc.). We define the union of two concepts as follows:
The core of the new concept is the union of the original cores. If it is not star-shaped, the same repair mechanism as for the intersection is used. We further set $\mu'_0 := \max(\mu_0^{(1)},\mu_0^{(2)})$ and compute $c'$ and $W'$ as described for the intersection.

\subsubsection{Subspace Projection.}
\label{CS:Operations:Projection}

Projecting a concept onto a subspace corresponds to focusing on certain domains while ignoring others. For instance, projecting the concept ``apple'' onto the color domain results in a property that describes the typical color of apples.
The projection of a concept $\widetilde{S}$ onto domains $\Delta_{S'} \subseteq \Delta_S$ is defined by projecting its core $S$ onto $\Delta_{S'}$, removing all weights that are irrelevant for $\Delta_{S'}$, and keeping $\mu_0$ and $c$ the same.

\subsubsection{Axis-Parallel Cut.}
\label{CS:Operations:Cut}

One can split a concept $\widetilde{S}$ into two parts (e.g., during a clustering process) by selecting a value $v$ on a dimension $d$ and by splitting each cuboid $C \in S$ into $C^{(+)} := \{x \in C \;|\; x_d \geq v\}$ and $C^{(-)} := \{x \in C \;|\; x_d \leq v\}$.\footnote{A strict inequality in the definition of $C^{(+)}$ or $C^{(-)}$ would not yield a cuboid.} It is easy to see that $S^{(+)} := \bigcup C_i^{(+)}$ and $S^{(-)} := \bigcup C_i^{(-)}$ are still valid cores. The parameters $\mu_0$, $c$, and $W$ remain unchanged when defining $\widetilde{S}^{(+)}$ and $\widetilde{S}^{(-)}$.

\subsubsection{Size.}
\label{CS:Operations:Size}

The size of a concept indicates its generality: Small concepts (like Granny Smith) tend to be more specific than larger concepts (e.g., apple). 
In \cite{Bechberger2017SGAI}, we derived the following formula for the size $M$ of a concept $\widetilde{S}$:\footnote{Where $m$ is the number of cuboids in $S$, $\delta(d)$ is the domain to which dimension $d$ belongs, $a_d := c \cdot w_{\delta(d)} \cdot \sqrt{w_d} \cdot (p^+_d - p^-_d)$, $\Delta_{\{d_1, \dots, d_i\}}$ is the domain structure that remains after removing from $\Delta$ all dimensions $d \notin \{d_1, \dots, d_i\}$, and $n_\delta := |\delta|$.}
\begin{align*}
M(\widetilde{S}) &= \sum_{l=1}^m \left((-1)^{l+1} \cdot \sum_{\substack{\{i_1,\dots,i_l\}\\\subseteq\{1,\dots,m\}}}M\left(\bigcap_{i \in \{i_1,\dots,i_l\}} \widetilde{C}_i\right)\right)\\
\text{with }M(\widetilde{C}) &= \frac{\mu_0}{c^n\prod_{d \in D} w_{\delta(d)} \sqrt{w_d}}
\sum_{i=0}^{n} \Bigg( 
\sum_{\substack{\{d_1,\dots,d_i\}\\ \subseteq D}} 
\left(\prod_{\substack{d \in \\ D \setminus \{d_1,\dots,d_i\}}} a_d\right) \cdot\\
&\hspace{5cm}\prod_{\substack{\delta \in\\ \Delta_{\{d_1,\dots,d_i\}}}} \left(
n_\delta! \cdot \frac{\pi^{\frac{n_\delta}{2}}}{\Gamma(\frac{n_\delta}{2}+1)}\right)\Bigg)
\end{align*}

\subsubsection{Subsethood.}
\label{CS:Operations:Subsethood}
The notion of subsethood gives rise to a concept hierarchy in a conceptual space: Because the region describing Granny Smith is a subset of the region describing apple, we know that Granny Smith is a specialization of the apple concept. We have defined a degree of subsethood as follows:
$$Sub(\widetilde{S}_1, \widetilde{S}_2) := \frac{M(\widetilde{S}_1 \cap \widetilde{S}_2)}{M(\widetilde{S}_1)}$$

As $\widetilde{S}_2$ sets the context for this subsethood judgement, we use $c^{(2)}$ and $W^{(2)}$ when computing the size of both the numerator and the denominator.

\subsubsection{Implication.}
\label{CS:Operations:Implication}

In order to support reasoning processes, an implication operation between different concepts is crucial. As it makes intuitive sense to consider $apple \Rightarrow red$ to be true to the degree to which $apple$ is a subset of $red$, we have defined the implication as follows:
$$Impl(\widetilde{S}_1, \widetilde{S}_2) := Sub(\widetilde{S}_1, \widetilde{S}_2)$$

\subsubsection{Betweenness.}
\label{CS:Operations:Betweenness}
We define the betweenness relation between concepts as the betweenness relation of the midpoints of their cores' central regions $P$.

\subsubsection{Similarity.}
\label{CS:Operations:Similarity}
We define the similarity relation of two concepts as the similarity relation of the midpoints of their cores' central regions $P$. The sensitivity parameter $c$ and the weights $W$ of the second concept (which sets the context of the comparison) are used to compute this similarity value.
\section{Implementation}
\label{Implementation}

\begin{figure}[t]
\centering
\includegraphics[width=\textwidth]{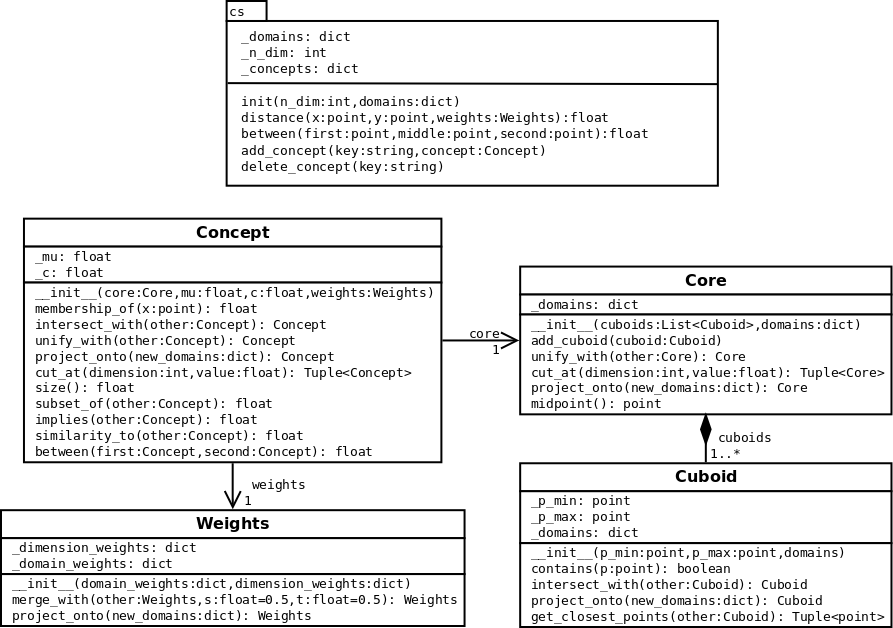}
\caption{Class diagram of our implementation.}
\label{fig:classDiagram}
\end{figure}
We have implemented our formalization in Python 2.7 and have made it publicly avaliable on GitHub\footnote{See \url{https://github.com/lbechberger/ConceptualSpaces/tree/v1.0.0}} \cite{Bechberger2017GitHub}. Figure \ref{fig:classDiagram} shows a class diagram illustrating the overall structure of our implementation. As one can see, each of the components from our definition (i.e., weights, cuboids, cores, and concepts) is represented by an individual class. Moreover, the ``cs'' module contains the overall domain structure of the conceptual space (represented as a dictionary mapping from domain identifiers to sets of dimensions) along with some utility functions (e.g., computing distance and betweenness of points). In order to define a new concept, one needs to use all of the classes, as all components of the concept need to be specified in detail. When operating with the concepts, it is however sufficient to use the \texttt{Concept} class which contains all the operations defined in Section \ref{CS:Operations}.

The implementation of most operations is rather straightforward. For instance, the subspace projection can be implemented by removing domains both from all cuboids of a concept and from its weights. The details about how these operations have been implemented are thus omitted from this paper. The intersection operation, however, has a more complex implementation and will be discussed in more detail. \\

\begin{figure}[tp]
\centering
\includegraphics[width=\textwidth]{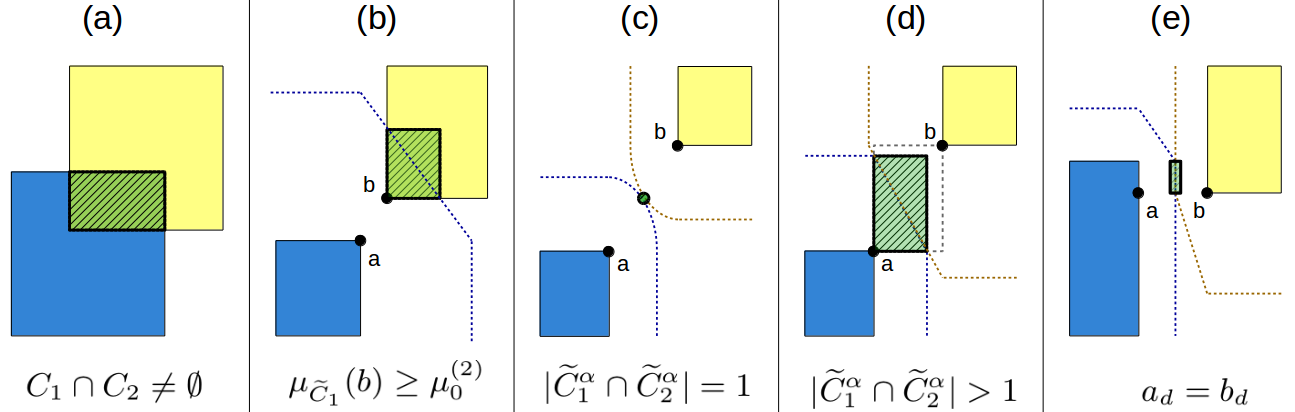}
\caption{Possible results of intersecting two fuzzy cuboids.}
\label{fig:IntersectionResults}
\end{figure}

The key challenge with respect to the intersection is to find the new core $S'$, i.e. the highest non-empty $\alpha$-cut intersection of the two sets. We simplify this problem by iterating over all combinations of cuboids $C_1 \in S_1, C_2 \in S_2$ and by looking at each pair of cuboids individually. In order to find the highest $\alpha$-cut intersection of the two cores, we simply take the maximum over all pairs of cuboids. Let $a \in C_1$ and $b \in C_2$ be the two closest points from the two cuboids under consideration (i.e., $\forall{x \in C_1, y \in C_2}: d(a,b) \leq d(x,y)$). When intersecting two fuzzified cuboids\footnote{The membership function $\mu_{\widetilde{C}}(x)$ can be obtained by replacing $S$ with $C$ in Def. \ref{def:FSSSS}.} $\widetilde{C}_1$ and $\widetilde{C}_2$, the following results are possible:
\begin{enumerate}
	\item The crisp cuboids have a nonempty intersection (Figure \ref{fig:IntersectionResults}a). In this case, we simply compute their crisp intersection.
	\item The $\mu_0$ parameters are different and the $\mu_0^{(i)}$-cut of $\widetilde{C}_j$ intersects with $C_i$ (Figure \ref{fig:IntersectionResults}b). In this case, we need to intersect $\widetilde{C}_j^{\mu_0^{(i)}}$ with $C_i$ and approximate the result by a cuboid. 
	\item The intersection of the two fuzzified cuboids consists of a single point $x^*$ lying between $a$ and $b$ (Figure \ref{fig:IntersectionResults}c). In this case, we define a trivial cuboid with $p^- = p^+ = x^*$.
	\item The intersection of the two fuzzified cuboids consists of a set of points (Figure \ref{fig:IntersectionResults}d). This can only happen if the $\alpha$-cut boundaries of both fuzzified cuboids are parallel to each other, which requires multiple domains to be involved and the weights of both concepts to be linearly dependent.
\end{enumerate}
Algorithm \ref{alg:intersection} shows how the intersection is implemented.
Lines 2 \& 3 cover the crisp intersection. After finding a pair of closest points $a,b$, we ignore from our further considerations all dimensions where $a_d = b_d$ (lines 5 \& 6). Lines 7-10 cover the second case. We use numerical optimization (from the \texttt{scipy.optimize} package) to find $x^*$ in line 12. If the weights are linearly dependent (line 14), we deal with the fourth case from above (lines 15-22): We look for points on the surface of the bounding box spanned by the points $a$ and $b$ that are both in $\widetilde{C}_1^{\alpha}$ and $\widetilde{C}_2^{\alpha}$. We iteratively look at the edges ($i=1$), faces ($i=2$), etc. of the bounding box until we find such points. We then approximate them with a cuboid. If the weights are not linearly dependent, we are in case 3 from above (line 24). Finally, in line 27 we extrude the identified cuboid in all dimensions where $a_d = b_d$ and where $a_d$ and $b_d$ can vary (cf. Figure \ref{fig:IntersectionResults}e).

\begin{algorithm}[t]
\caption{Highest non-empty $\alpha$-cut of two cuboids}
\label{alg:intersection}
\begin{algorithmic}[1]
\Function{intersect\_cuboids}{Cuboid $C_1$, Cuboid $C_2$, Concept $\widetilde{S}_1$, Concept $\widetilde{S}_2$}

\If{$C_1 \cap C_2 \neq \emptyset$}
    \State $\alpha \gets \min(\mu_0^{(1)}, \mu_0^{(2)})$, $C \gets C_1 \cap C_2$
\Else
    \State Find closest points $a \in C_1, b \in C_2$
    \State Ignore all dimensions $d$ where $a_d = b_d$ 
    \If{$\mu_{\widetilde{C}_1}(b) \geq \mu_0^{(2)}$ }
        \State $\alpha \gets \mu_0^{(2)}$, $C \gets \text{cuboid approximation of }\widetilde{C}_1^{\mu_0^{(2)}} \cap C_2$
    \ElsIf{$\mu_{\widetilde{C}_2}(a) \geq \mu_0^{(1)}$}
    	\State $\alpha \gets \mu_0^{(1)}$, $C \gets \text{cuboid approximation of }\widetilde{C}_2^{\mu_0^{(1)}} \cap C_1$
    \Else
    	\State Find $x^* = \argmax_{x \in CS}(\mu_{\widetilde{C}_1}(x))$ with $\mu_{\widetilde{C}_1}(x) = \mu_{\widetilde{C}_2}(x)$
    	\State $\alpha \gets \mu_{\widetilde{C}_1}(x^*)$
    	\If{$\exists {t \in \bbbr}: \forall{d\text{ with }a_d \neq b_d}: w^{(1)}_{\delta(d)} \cdot \sqrt{w^{(1)}_{d}} = t \cdot w^{(2)}_{\delta(d)} \cdot \sqrt{w^{(2)}_{d}}$}
			\For{$i = 1$ to $|\{d: a_d \neq b_d\}| - 1$}
				\State Find all $x$ on the $i$-faces of the bounding box spanned by $a$ and $b$
				\State \hspace{1cm} with $\mu_{\widetilde{C}_1}(x) = \mu_{\widetilde{C}_2}(x) = \alpha$
				\If {found at least one $x$}
					\State $C \gets$ cuboid-approximation of the set of all $x$
					\State \textbf{break}
				\EndIf
			\EndFor
    	\Else
    		\State $C \gets$ trivial cuboid consisting of $x^*$
    	\EndIf
    \EndIf
    \State Extrude $C$ in all dimensions $d$ where $a_d = b_d$
\EndIf
\State \Return $\alpha, C$
\EndFunction
\end{algorithmic}
\end{algorithm}

\section{Example}
\label{Examples}

\begin{figure}[tp]
\centering
\includegraphics[width=0.85\textwidth]{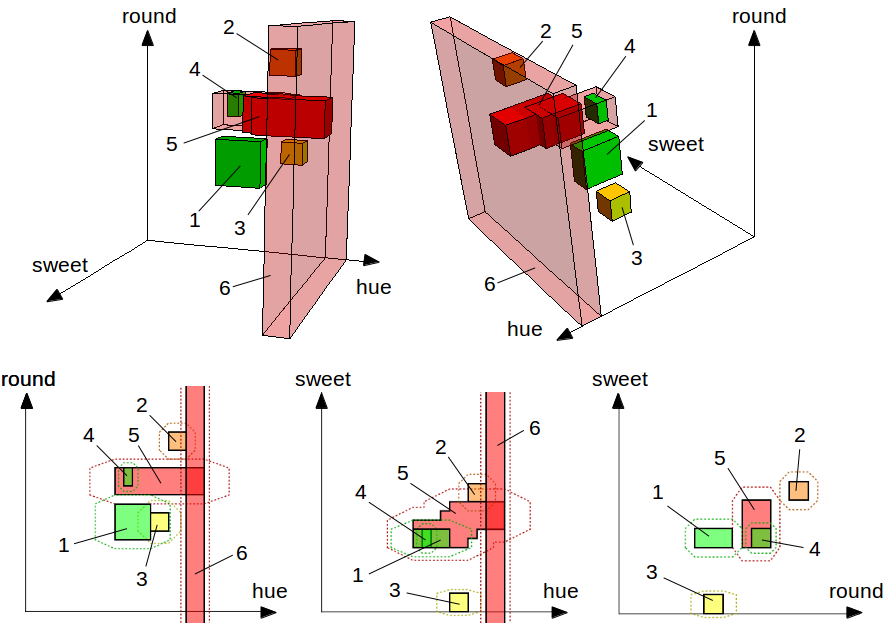}
\caption{Top: Three-dimensional fruit space (only cores). Bottom: Two-dimensional visualization of the fruit space (cores and 0.5-cuts). The concepts are labeled as follows: pear (1), orange (2), lemon (3), Granny Smith (4), apple (5), red (6).}
\label{fig:FruitSpace}
\end{figure}

Our implementation of the conceptual spaces framework contains a simple toy example -- a three-dimensional conceptual space for fruits, defined as follows:
$$\Delta = \{\delta_{color} = \{d_{hue}\},\delta_{shape} = \{d_{round}\},\delta_{taste} = \{d_{sweet}\}\}$$
$d_{hue}$ describes the hue of the observation's color, ranging from $0.00$ (purple) to $1.00$ (red). $d_{round}$ measures the percentage to which the bounding circle of an object is filled. $d_{sweet}$ represents the relative amount of sugar contained in the fruit, ranging from 0.00 (no sugar) to 1.00 (high sugar content). As all domains are one-dimensional, the dimension weights $w_{d}$ are always equal to 1.00 for all concepts. We assume that the dimensions are ordered like this: $d_{hue},d_{round},d_{sweet}$.
Table \ref{tab:FruitSpace} defines some concepts in this space\footnote{Due to space restrictions, we only show a subset of the concepts defined in the demo.} and Figure \ref{fig:FruitSpace} visualizes them.
\begin{table}[t]
  \centering
  \begin{tabular}{|l||c|c|c|c|c|c|c|c|}
    \hline
    Concept 	& $\Delta_S$& $p^-$ 				& $p^+$ 				& $\mu_0$ 	& $c$ 	& \multicolumn{3}{|c|}{$W$}\\ 
    & & & & & & $w_{\delta_{color}}$ & $w_{\delta_{shape}}$ & $w_{\delta_{taste}}$\\ \hline \hline
    Pear		& $\Delta$	& (0.50, 0.40, 0.35)	& (0.70, 0.60, 0.45)	& 1.0		& 12.0	& 0.50 & 1.25 & 1.25 \\ \hline
    Orange		& $\Delta$	& (0.80, 0.90, 0.60)	& (0.90, 1.00, 0.70)	& 1.0		& 15.0	& 1.00 & 1.00 & 1.00 \\ \hline
    Lemon		& $\Delta$	& (0.70, 0.45, 0.00)	& (0.80, 0.55, 0.10)	& 1.0		& 20.0	& 0.50 & 0.50 & 2.00 \\ \hline
    Granny & \multirow{2}{*}{$\Delta$}	& \multirow{2}{*}{(0.55, 0.70, 0.35)}	& \multirow{2}{*}{(0.60, 0.80, 0.45)}	& \multirow{2}{*}{1.0}		& \multirow{2}{*}{25.0} & \multirow{2}{*}{1.00} & \multirow{2}{*}{1.00} &  \multirow{2}{*}{1.00}\\ 
    Smith & & & & & & & &\\ \hline
    \multirow{3}{*}{Apple}	& \multirow{3}{*}{$\Delta$}	& (0.50, 0.65, 0.35)	& (0.80, 0.80, 0.50)	& \multirow{3}{*}{1.0}		& \multirow{3}{*}{10.0}	& \multirow{3}{*}{0.50} & \multirow{3}{*}{1.50} &  \multirow{3}{*}{1.00} \\ 
    			&			& (0.65, 0.65, 0.40)	& (0.85, 0.80, 0.55)	&			&		&  & & \\ 
    			&			& (0.70, 0.65, 0.45)	& (1.00, 0.80, 0.60)	&			&		&  & & \\ \hline
    Red			& $\{\delta_{color}\}$ & (0.90, -$\infty$, -$\infty$) & (1.00, +$\infty$, +$\infty$) & 1.0 & 20.0 & 1.00 & -- & -- \\ \hline
  \end{tabular}\\[1ex]
  \caption{Definitions of several concepts.}
  \label{tab:FruitSpace}
\end{table}
The conceptual space is defined as follows in the code:
\begin{lstlisting}[language=Python]
domains = {'color':[0], 'shape':[1], 'taste':[2]}
space.init(3, domains)
\end{lstlisting}
Concepts can be defined as follows:
\begin{lstlisting}[language=Python]
c_pear = Cuboid([0.5, 0.4, 0.35], [0.7, 0.6, 0.45], domains)
s_pear = Core([c_pear], domains)
w_pear = Weights({'color':0.50, 'shape':1.25, 'taste':1.25}, {'color':{0:1.0}, 'shape':{1:1.0}, 'taste':{2:1.0}})
pear = Concept(s_pear, 1.0, 12.0, w_pear)
\end{lstlisting}

We can load the definition of this fruit space into our python interpreter and apply the different operations described in Section \ref{CS} to these concepts. This looks for example as follows:
\begin{lstlisting}[language=Python]
>>> execfile('fruit_space.py')
>>> granny_smith.subset_of(apple)
1.0
>>> apple.implies(red)
0.3333333333333332
>>> apple.between(lemon, orange)
1.0
>>> pear.similarity_to(apple)
0.007635094218859955
>>> pear.similarity_to(lemon)
1.8553913626159717e-07
>>> print apple.intersect_with(pear)
core: {[0.5, 0.625, 0.35]-[0.7, 0.625, 0.45]}
mu: 0.6872892788
c: 10.0
weights: <{'color': 0.5, 'taste': 1.125, 'shape': 1.375},
   {'color': {0: 1.0}, 'taste': {2: 1.0}, 'shape': {1: 1.0}}>
>>> apple.size(), pear.size()
(0.10483333333333335, 0.041481481481481466)
>>> (apple.unify_with(pear)).size()
0.146900844381
>>> print lemon.project_onto({"color":[0]})
core: {[0.7, -inf, -inf]-[0.8, inf, inf]}
mu: 1.0
c: 20.0
weights: <{'color': 1.0},{'color': {0: 1.0}}>
\end{lstlisting}

\section{Related Work}
\label{RelatedWork}

Our work is of course not the first attempt to devise an implementable formalization of the conceptual spaces framework.

An early and very thorough formalization was done by Aisbett \& Gibbon \cite{Aisbett2001}. Like we, they consider concepts to be regions in the overall conceptual space. However, they stick with G\"{a}rdenfors' assumption of convexity and do not define concepts in a parametric way. Their formalization targets the interplay of symbols and geometric representations, but it is too abstract to be implementable. 

Rickard \cite{Rickard2006} provides a formalization based on fuzziness. He represents concepts as co-occurence matrices of their properties. By using some mathematical transformations, he interprets these matrices as fuzzy sets on the universe of ordered property pairs. As properties and concepts are represented in different ways, one has to use different learning and reasoning mechanisms for them. Rickard's formalization is also not easy to work with due to the complex mathematical transformations involved.

Adams \& Raubal \cite{Adams2009} represent concepts by one convex polytope per domain. This allows for efficient computations while being potentially more expressive than our cuboid-based representation. However, correlations between different domains are not taken into account. Adams \& Raubal also define operations on concepts, namely intersection, similarity computation, and concept combination. This makes their formalization quite similar in spirit to ours.

Lewis \& Lawry \cite{Lewis2016} formalize conceptual spaces using random set theory. They define properties as random sets within single domains, and concepts as random sets in a boolean space whose dimensions indicate the presence or absence of properties. Their approach is similar to ours in using a distance-based membership function to a set of prototypical points. However, their work purely focuses on modeling conjunctive concept combinations and does not consider correlations between domains. 

To the best of our knowledge, none of the above mentioned formalizations have a publicly accessible implementation.
In \cite{Lieto2017}, Lieto et al. present a hybrid architecture that represents concepts by using both description logics and conceptual spaces. This way, symbolic ontological information and similarity-based ``common sense'' knowledge can be used in an integrated way. Each concept is represented in the conceptual space by a single prototypical point and a number of exemplar points. Correlations between domains can therefore only be encoded through the selection of appropriate exemplars. Their work focuses on classification tasks and does therefore not provide any operations for combining different concepts. With respect to the larger number of supported operations, our implementation can therefore be considered more general than theirs. The implementation of their system\footnote{See \url{http://www.dualpeccs.di.unito.it/download.html}.} is the only publicly available implementation of the conceptual spaces framework we are currently aware of. In contrast to our work, it however comes without any publicly available source code\footnote{The source code of an earlier and more limited version of their system can be found here: \url{http://www.di.unito.it/~lieto/cc_classifier.html}.}.

\section{Conclusion and Future Work}
\label{Conclusion}

In this paper, we presented a comprehensive implementation of the conceptual spaces framework. This implementation and its source code are publicly avaliable and can be used by any researcher interested in conceptual spaces. We think that our implementation can be a good foundation for practical research on conceptual spaces and that it will considerably facilitate research in this area.

In future work, we will implement a visualization toolbox in order to enrich the presented implementation with visual output. Moreover, we will use this implementation to apply machine learning algorithms in conceptual spaces. Needless to say, any future extensions of our formalization will also be incorporated into future versions of this implementation.

\bibliographystyle{splncs03}
\bibliography{/home/lbechberger/Documents/Papers/jabref.bib}

\end{document}